\documentclass[10pt,twocolumn,letterpaper]{article}

\usepackage{cvpr}              % To produce the CAMERA-READY version
\usepackage{graphicx}
\usepackage{amsmath}
\usepackage{amssymb}
\usepackage{booktabs}
\usepackage[numbers]{natbib}
\usepackage{mathtools}
\usepackage{amsthm}

\usepackage[pagebackref,breaklinks,colorlinks]{hyperref}

\usepackage[capitalize]{cleveref}
\crefname{section}{Sec.}{Secs.}
\Crefname{section}{Section}{Sections}
\Crefname{table}{Table}{Tables}
\crefname{table}{Tab.}{Tabs.}

\newcommand{\algo}{Adaptive Distillation of Adapters~}
\newcommand{\al}{ADA}

\def\cvprPaperID{*****} % *** Enter the CVPR Paper ID here
\def\confName{CVPR}
\def\confYear{2022}

\begin{document}

\title{Continual Learning with Transformers for Image Classification}

\author{Beyza Ermis\\
AWS, Berlin\\
{\tt\small ermibeyz@amazon.com}
\and
Giovanni Zappella\\
AWS, Berlin\\
{\tt\small zappella@amazon.com}
\and
Martin Wistuba\\
AWS, Berlin\\
{\tt\small marwistu@amazon.com}
\and
Aditya Rawal\\
AWS, Santa Clara\\
{\tt\small adirawal@amazon.com}
\and
C\'edric Archambeau\\
AWS, Berlin\\
{\tt\small cedrica@amazon.com}
}
\maketitle

%%%%%%%%% ABSTRACT
\begin{abstract}

In many real-world scenarios, data to train machine learning models become available over time.
However, neural network models struggle to continually learn new concepts without forgetting what has been learnt in the past.
This phenomenon is known as \textit{catastrophic forgetting} and it is often difficult to prevent due to practical
constraints, such as the amount of data that can be stored or the limited computation sources that can be used. 
Moreover, training large neural networks, such as Transformers, from scratch is very costly and requires a vast amount of training data,
which might not be available in the application domain of interest.
A recent trend indicates that dynamic architectures based on an expansion of the parameters can reduce catastrophic
forgetting efficiently in continual learning, but this needs complex tuning to balance the growing number of parameters 
and barely share any information across tasks. As a result, they struggle to scale to a large number of tasks without significant overhead.
In this paper, we validate in the computer vision domain a recent solution called Adaptive Distillation of Adapters (ADA),
which is developed to perform continual learning using pre-trained Transformers and Adapters on text classification tasks.
We empirically demonstrate on different classification tasks that this method maintains a good predictive performance without retraining the model or
increasing the number of model parameters over the time.
Besides it is significantly faster at inference time compared to the state-of-the-art methods.

\end{abstract}

%%%%%%%%% BODY TEXT
\section{Introduction}
\label{sec:intro}

The ability to learn from evolving streams of training data is important for many real-world applications.
While neural networks showed a great ability to learn a task, but when confronted with a sequence of different ones
they tend to override the previous concepts. Deep networks suffer heavily from this phenomenon called \textit{catastrophic forgetting}
(CF)~\cite{mccloskey1989catastrophic}, impeding continual or lifelong learning.
A growing amount of efforts have emerged to tackle catastrophic 
forgetting~\cite{douillard2020podnet,hou2019learning,kirkpatrick2017overcoming,rebuffi2017icarl,wu2019large,yan2021dynamically}
in continual learning (CL).
Existing methods can be roughly categorized as replay-based 
methods~\cite{lopez2017gradient,rolnick2018experience,d2019episodic,chaudhry2019continual,wang2020efficient} 
that retain some training data of old tasks and use them in learning a new task to circumvent the issue of CF; 
regularization-based methods~\cite{kirkpatrick2017overcoming,aljundi2018selfless,huang2021continual} add 
a regularization term to the loss to consolidate previous knowledge when
learning a new task; and parameter isolation methods (PIM) that can dynamically expand 
the network architectures~\cite{yoon2017lifelong,li2019learn} or re-arrange their structures~\cite{golkar2019continual,hung2019compacting}.
Replay-based methods explicitly retrain on a subset of stored samples while training 
on new tasks. That can induce dramatic memory overhead when tackling a
large number of tasks or regulatory-related issues due to the data storage. 
Regularization-based methods are
often difficult to tune due to the sensitivity to the importance of the
regularization term. In practice, this complexity often leads to poor performance. 
Parameter-isolation methods either keep augmenting additional parameters 
with the consequence of significantly increasing the number of parameters,
or need complex pruning as post-processing with requirement to know which parameters should be kept/pruned. 

Inspired by the significant achievement of Transformers~\cite{vaswani2017attention} in NLP, some pioneering works have recently been done 
on adapting transformer architectures to Computer Vision (CV).
Vision Transformer (ViT)~\cite{dosovitskiy2020image} showed that a pure 
Transformer applied directly to a sequence of image patches can perform well
on image classification tasks if the training dataset is sufficiently large.
Data-Efficient Image Transformers (DeiT)~\cite{touvron2021training} further demonstrated that Transformers
can be successfully trained on standard datasets, such as ImageNet-1K~\cite{deng2009imagenet}. 
Besides, some recent studies~\cite{yin2019benchmarking,brown2020language,li2022technical} 
showed transformers have a good ability to generalize well to
new domains with a few samples. 
To this end, \emph{we leverage the vision transformers} to improve the ease of use of CL frameworks for real-world applications.
To the best of our knowledge, only a few recent works~\cite{li2022technical,douillard2021dytox} 
have applied the transformer architecture to CL on image datasets,
but they also require rehearsal memory or training a new transformer model from scratch (cannot work with pre-trained transformer models).

Due to the drawbacks that we listed above, existing methods do not seem to fit 
the requirements of many practical CL applications. We found the work of \cite{ermis2022ada}
as a promising candidate to adopt since it does not require re-training the 
model nor storing the old task data, keep the memory and resources consumption limited 
(e.g. constant number of parameters) as the number of tasks grows and can work with pre-trained transformer models.
This method stores a small set of covariates to be used for distillation, 
but given the limited size and usage, they can easily be replaced with external data sources 
(e.g., data provided by the owner of the model and not subject to the same legal restrictions of user data). 
It introduces a reasonable amount of additional parameters
(about 12\% of the BERT size in their experiments) and only adds a new linear head for every task.
Moreover, it is tested with several pre-trained transformer models for text (i.e., BERT, DistillBERT and RoBERTa).
Nevertheless, there is no evidence that this may work for CV Transformers. Our main focus is to answer the question: 
Can this technique provide good performance also for CV?

In this study, we address an image classification problem on a sequence of classification tasks provided in sequence using pre-trained models: ViT and DeiT. 
While the solution can be used in several different tasks, we limit our study to image classification since image classification is core to computer vision
and it is often used as a benchmark to measure progress in image understanding. We conjecture that the behaviour observed in these experiments will be
informative also for related tasks such as detection or segmentation.

We test the efficiency of \algo (\al) on CIFAR100 and MiniImageNet. Our main contributions are: 1) using Adapters approach with vision
transformers for the first time on continual image classification tasks 2) validate that Adapters work with vision transformers and show that \algo (\al)
can achieve predictive performance on-par with memory-hungry methods such AdapterFusion~\cite{pfeiffer-etal-2021-adapterfusion}, 3) adding a benchmark with 
CL methods such Elastic Weight Consolidation (EWC)~\cite{kirkpatrick2017overcoming} and Experience Replay (ER)~\cite{rolnick2018experience}
by using transformer architectures.

\paragraph{Related Work.} We discuss the related work in two folds: Adapters and CL with vision transformers.
Residual adapters~\cite{rebuffi2018efficient} adds a few learnable residual layers to the standard ResNet model and
train only these newly added layers, has been proposed as a transfer learning technique used for multi-domain learning and domain adaptation.
These adapters are designed specifically for ResNet models, so they are not generalizable to other networks. 
Recently Houlsby \emph{et al.}~\cite{houlsby2019parameter} proposed Adapters for Transformers in natural language processing (NLP) to 
isolate task-specific knowledge for multi-task learning. However, in the sequential learning setting, Adapters keep increasing the model
parameters with each task, so the memory requirements.
In the CL direction, \cite{ermis2022ada} uses Adapters for text classification, 
but by using distillation after training a set of task, it manages to keep the 
memory size constant while maintaining a good prediction performance for both old and new tasks. 
Benchmarking this approach on CV tasks is the main focus of our work.

Two recent works are related with the method we are testing. In~\cite{li2022technical}, 
for each task, before training on a new task, the model is copied and
fixed to be used as the teacher model in the distillation phase. 
The student model is trained on both new task samples together with the knowledge distillation loss
that uses samples from old tasks which is stored in the rehearsal memory. 
In~\cite{douillard2021dytox}, they aim to learn a unified model that will classify an increasingly
growing number of classes by building upon a new architecture. However, they need to train a new transformer
model, where the process is very costly. Our main goal is to use public pre-trained models.

\section{Problem setup}
\label{sec:setup}

In this section we formalize our goal of CL on a sequence of image classification tasks 
$\lbrace T_1, \dots, T_N \rbrace$ where each task $T_i$ contains
a different set of image-label training pairs $(x_{1:t}^i, y_{1:t}^i)$.
Each task $T_i$ may contain $c$ new classes namely $Y_i=\lbrace Y_i^1, \dots, Y_i^c\rbrace$
and each new class has $t$ examples. $T_i$ is sampled iid from a distribution $D_i(X_i, Y_i)$.
Each task represents a different classification problem and the learner creates a new classification head for each.
The task identifier is provided also at inference time.

The goal of the learner is to learn a set of parameters $\tilde{\Theta}$ such that 
$\frac{1}{N} \sum_{i \in \{1,\dots, N\}} \text{loss}(T_i; \tilde{\Theta})$ is minimized.
In our specific case, $\tilde{\Theta}$ is composed of a set of parameters $\Theta$
provided by a pre-trained model and, depending on the algorithm, additional network parts whose
weights need to be learnt. In its simplest case this additional set of model parameters can just be a head model but,
some algorithms use significantly more elaborate functions.

For the training of task $T_i$, the algorithm can access the data provided in the current batch and some data which eventually
stored in memory. To evaluate the system, the test data consists of examples across all the previous tasks. All methods in the following
sections receive as input a pre-trained language model $f_{\Theta}(.)$, e.g., ViT~\cite{dosovitskiy2020image}, parameterized by $\Theta$.
The pre-trained model receives an input image, called $x_i$ and it is able to compute a representation for it.

\section{The ADA algorithm}
\label{sec:ada}

In this section we provide an overview of the main components used by \algo (\al) and the algorithm itself.
ADA is performed in two steps: the first step is to train an adapter model and a new classifier using the new 
task $T_i$'s training dataset $D_i$, which is referred as the \emph{new} model; the second step is to consolidate the
\emph{old} model(s), the model(s) obtained in the previous round, and new model.
 
\begin{figure}[ht!]
\hspace*{-2mm}
\begin{minipage}{0.15\textwidth}
\centering
\includegraphics[scale=0.15]{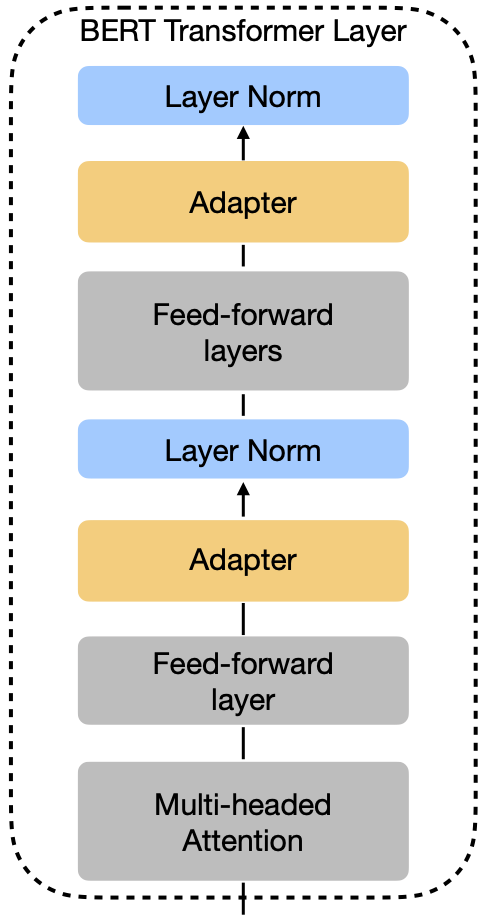}
\end{minipage}
\begin{minipage}{0.15\textwidth}
\centering
\includegraphics[scale=0.14]{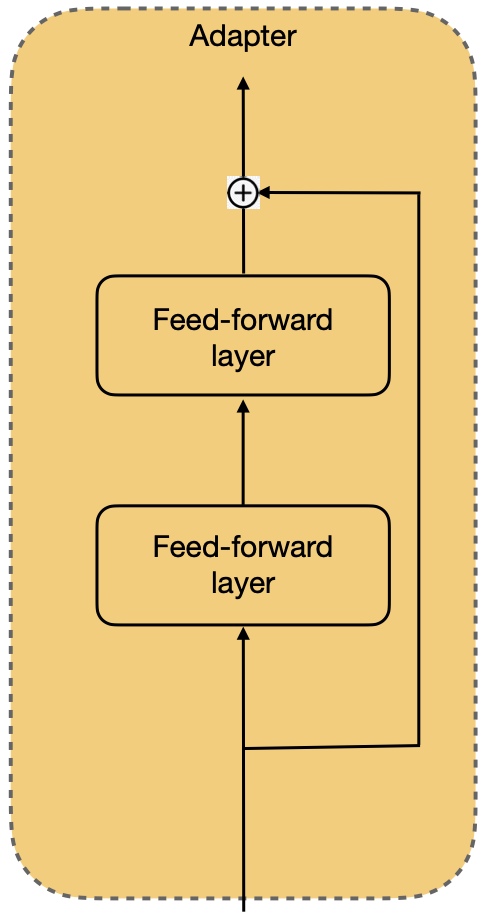}
\end{minipage}
\hspace*{1mm}
\begin{minipage}{0.16\textwidth}
\centering
\includegraphics[scale=0.15]{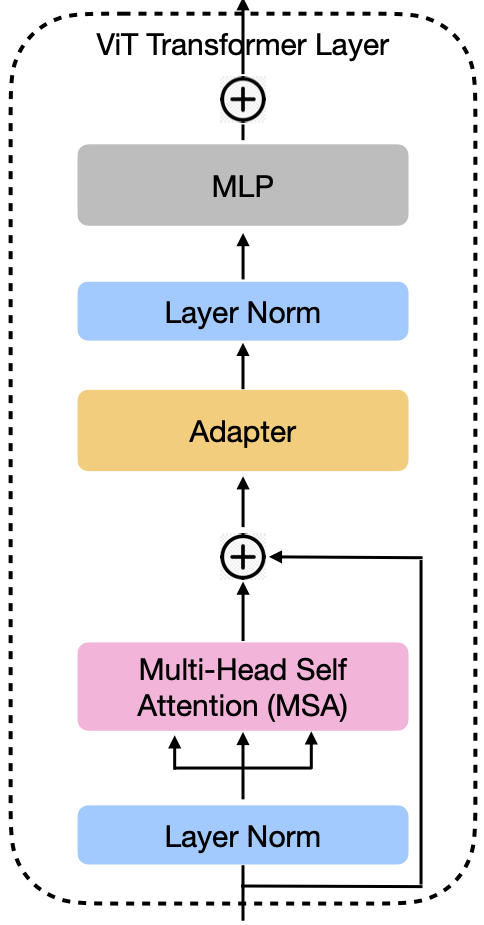}
\end{minipage}

\caption{\emph{Left} shows Adapter-BERT~\cite{houlsby2019parameter} in a BERT transformer layer, and \emph{Middle} shows the adapter architecture. Depending on configuration (Houlsby~\cite{houlsby2019parameter} or Pfeiffer~\cite{pfeiffer-etal-2021-adapterfusion}), only the top Adapter can be used. Right shows our Adapter implementation in a ViT transformer layer.
As in Adapter-BERT, we added an adapter before layer norm and feed-forward layers (MLP).}
\label{fig:Adapters}
\end{figure}

\paragraph{Adapters.} Adapters share a large set of parameters $\Theta$ across all tasks and introduce a small number of task-specific parameters $\Phi_i$. 
Current work on adapters focuses on training adapters for each task separately. For each of the $N$ tasks, the model is 
initialized with parameters of a pre-trained model $\Theta$. 
In addition, a set of new and randomly initialized adapter
parameters $\Phi_i$ are introduced for tasks $i \in \lbrace 1, \dots, N \rbrace$.
The parameters $\Theta$ are fixed and only the parameters $\Phi_i$ are trained. 
This makes it possible to train adapters for all $N$ tasks, and store the corresponding 
knowledge in designated parts of the model. 
The objective for each task $i \in \lbrace 1, \dots, N \rbrace$ is of the form:
$\Phi_i \leftarrow \arg\min_\Phi L_i(D_i;\Theta,\Phi)$. \\ 
In this work, we propose to define and use adapters for vision transformers~\cite{dosovitskiy2020image,touvron2021training} 
While this provides good predictive performance, in the CL setting, new tasks are added 
sequentially and storing a large set of adapters $\Phi_1, \dots, \Phi_N$ is practically infeasible.
As in AdapterBERT, we insert a 2-layer fully-connected network in each transformer layer of ViT and DeiT (see Figure~\ref{fig:Adapters}).
DeiT is built upon the ViT architecture, so the Adapter is added in the same way.

\paragraph{Distillation of Adapters.} For each new task $T_i$, the adapter parameters $\Phi_i$ are added to the model, while the pre-trained 
model parameters $\Theta$ are kept frozen. 
Only the task-specific model parameters $\Phi_i$ and the head model parameters $h_i$ are trained for the current task. The model $f_i(x;\Theta,\Phi_i)$,
with parameters $\Theta$ and $\Phi_i$ is called the \emph{new} model $f_{new}$. 
When a prediction for $T_i$ is required the corresponding head model $h_i$ is called.
The distillation of the two models has the following objective:
\[
    f(x;\Theta,\Phi_c) = 
\begin{cases}
    f_{old}(x;\Theta,\Phi_{i},h_i)[i], & 1 \leq i \leq n-1 \\
    f_{new}(x;\Theta,\Phi_n,h_n)[i], & i=n
\end{cases}
\]
where $i$ denotes the index of the considered task and $f_{old}$ is the model trained on the previous tasks.
The output of the consolidated model approximates the combination of the
model outputs of the old model and the new model. To achieve this,
the outputs of the old model and the new model are employed as supervisory signals in joint training of
the consolidated model $\Phi_c$.\\
To this purpose, they use the double distillation loss proposed by \cite{zhang2020class} 
to train a new adapter that is used with the pre-trained model to classify both old and newly 
learned tasks. 
The distillation process proceeds as follows: $f_{old}$ and $f_{new}$ are freezed, 
run a feed-forward pass for each training sample, and collect the \emph{logits} 
of the two models 
$\hat{y}_{old}=\left[\hat{y}^1, \dots, \hat{y}^{n-1}\right]$ and $\hat{y}_{new}=\hat{y}^n$ respectively,
where the super-script is the class label associated with the
neuron in the model.
Then the difference between the logits produced by the consolidated model and the 
combination of logits generated by the two existing specialist models
based on $L_2$-loss is minimized:
\begin{align}
L_d(y,\hat{y}) = \frac{1}{n}\sum_{j=1}^t \big(y^j-\hat{y}^j\big)^2 ,
\label{eq:distillation_loss}
\end{align}
where $y^j$ are the \emph{logits} produced by the consolidated model for the $i^{th}$ 
task and $\hat{y}$ is the concatenation of $\hat{y}_{old}$ and $\hat{y}_{new}$. 
The training objective for consolidation is given by
\begin{align}
\min_{\Theta,\Phi_c} \frac{1}{|\mathcal{U}|} \sum_{x_j \in \mathcal{U}} L_d(y,\hat{y}) ,
\label{eq:consolidation_objective}
\end{align}
where $\mathcal{U}$ denotes the unlabeled training data used for distillation. 
After the consolidation, the adapter parameters $\Phi_c$ are used for $f_{old}$ in the next round.

\paragraph{Transferability Estimation.} ADA keeps a pool of adapters and the selection of the adapter to be distilled is based on transferability. 
The intuition behind this choice is that highly similar tasks will interfere less with each other and so will cause significantly less forgetting. 
In~\cite{ermis2022ada} they leverage two common methods for transferability estimation: 
(1) \emph{Log Expected Empirical Prediction} (LEEP)~\cite{nguyen2020leep} and 
(2) TransRate~\cite{huang2021frustratingly}. \\
\emph{LEEP} is a measure (or a score) that can tell us, without training on the target data set, how effectively the transfer learning algorithms can transfer knowledge
learned in the source model $\Theta_s$ to the target task, using the target data set $\mathcal{D}$. The details of LEEP can be checked in~\cite{nguyen2020leep} and \cite{ermis2022ada}.

\noindent\emph{TransRate} measures the transferability as the mutual information between the features of target examples extracted by a pre-trained model and labels of them with
a single pass through the target data. It achieves minimal value when the data covariance matrices of all classes are the same. 
In this case, it is impossible to separate the data from different classes and no classifier can perform better than random guesses. 
To see the details how the knowledge transfer from a source task to a target task is measured, please see~\cite{huang2021frustratingly}.

\paragraph{Algorithm.} ADA procedure is detailed in Algorithm~\ref{algo:ADA}.
For every new task the algorithm trains a new adapter and head model (called $\Phi_n$ and $h_n$).
If the adapters pool did not reach the maximum size yet (controlled by $K$), it just adds it to the pool.
If the pool reached the maximum size, the algorithm is forced to select one of the adapters already
in the pool and distill it together with the newly trained one.
In order to select which adapter to distill it leverages the transferability scores (e.g., LEEP or TransRate).
Once the adapter in the pool with the highest transferability score (called $f_{old}^{j^{\ast}}$) is identified,
it consolidates that adapter and the newly trained one into a new adapter and replaces the old one present in the pool.
In order to be able to make effective predictions, the algorithm also keeps a mapping (in the map $m$) of which adapter in the pool
must be used in combination with each of the task-specific heads. 

\begin{algorithm}[ht]
\small
\caption{Adaptive Distillation of Adapters (ADA)}
\begin{algorithmic}[1]
%\REQUIRE $\lbrace T_1, \dots, T_n\rbrace$, $\Theta$, $K$, adapter configuration
\REQUIRE $\Theta$: pre-trained model, $K$: adapters pool size
\STATE Freeze $\Theta$
\STATE Create $m = Map()$

\FOR{$n \gets 1$ to $N$}
\STATE A task $T_n$ is received
\STATE Initialize $\Phi_n$
\STATE Process $T_n$ and train new model $f_n(x;\Theta,\Phi_n)$ and head $h_n$ 
\STATE Sample from $T_n$ and add to distillation data $\mathcal{D}_{distill}$
\IF {$n \leq K$}
\STATE Set $f_n(x;\Theta,\Phi_1)$ to $f_{old}^n$
\ELSE
\STATE $j^{\ast} \gets \arg\max_{j \in \{1, \dots, K\}} \textit{TRANSCORE}(T_n,f_n,f_{old}^j)$
\STATE Add ($n, j^{\ast}$) to $m$
\STATE Consolidate model: \\
\hspace*{5mm} $f_{old}^{j^{\ast}}$ = \textit{DISTILLATION}($f_{old}^{j^{\ast}}$, $f_n$, $\mathcal{D}_{distill}$)
\ENDIF
\STATE Serve predictions for any task $i \leq n$ using $h_i$ and $f_{old}^{m(i)}$
\ENDFOR
\vspace*{3mm}
%\STATE // Model consolidation with old and new model
\STATE \textit{DISTILLATION}($f_{old}$, $f_n$, $\mathcal{D}_{distill}$):
\STATE \hspace*{2mm} Get soft targets $\hat{y}_{old}$ from old model $f_{old}$ with $\mathcal{D}_{distill}$
\STATE \hspace*{2mm} Get soft targets $\hat{y}_{new}$ from new model $f_n$ with $\mathcal{D}_{distill}$
\STATE \hspace*{2mm} Initialize $\Phi_c$
\STATE \hspace*{2mm} Compute distillation loss as in Equation (\ref{eq:distillation_loss}) and train model $f(x;\Theta,\Phi_c)$
\STATE \hspace*{2mm} \textbf{return} $f$
\end{algorithmic}
\label{algo:ADA}
\end{algorithm}
\normalsize

% the following doesn't seem very useful, it's already said
\iffalse
\paragraph{Differences with other Adapters-based methods.} Current work on adapters focuses on training adapters for each task separately. 
\al keeps a fixed amount of adapters in memory and takes transferability of representations into account to effectively
consolidate newly created adapters with previously created ones with unlabeled training data which is required for consolidation.
For \al, we have to fix a budget for the number of adapters $K$ that we can keep, for instance due to memory constraints, 
and the algorithm selects the adapter model from the pool of $K$ adapters to consolidate by
using scores that are computed based on transferability estimation.
\fi

\section{Experiments}
\label{sec:exp}

As discussed in Section~\ref{sec:intro}, we would like to test ADA due to three main characteristics:
i) limited or no data (external or synthetic data can be used for distillation) storage required;
ii) almost-constant number of parameters added with the growing number of tasks and 
iii) interoperability with pre-trained Transformers available in public repositories.
To this purpose, we designed a set of experiments on CIFAR100~\cite{krizhevsky2009learning}
and MiniImageNet~\cite{russakovsky2015imagenet}.
The complete setup is described in the next sections.
%
%% Figure1. Accuracy comparison of methods
\begin{figure*}[ht!]
\centering
\begin{minipage}{0.45\textwidth}
\includegraphics[width=\textwidth]{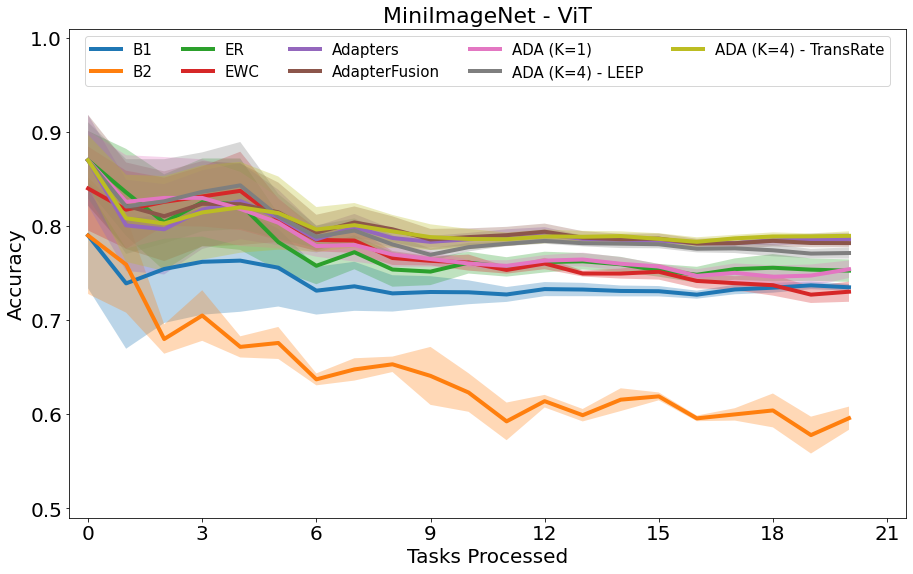}
\end{minipage}
\begin{minipage}{0.45\textwidth}
\includegraphics[width=\textwidth]{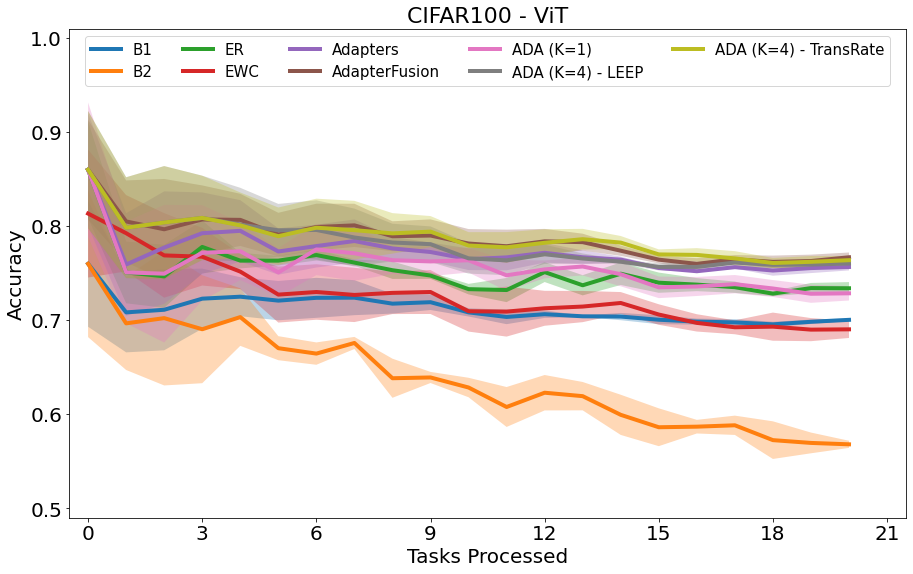}
\end{minipage}
\\
\centering
\begin{minipage}{0.45\textwidth}
\includegraphics[width=\textwidth]{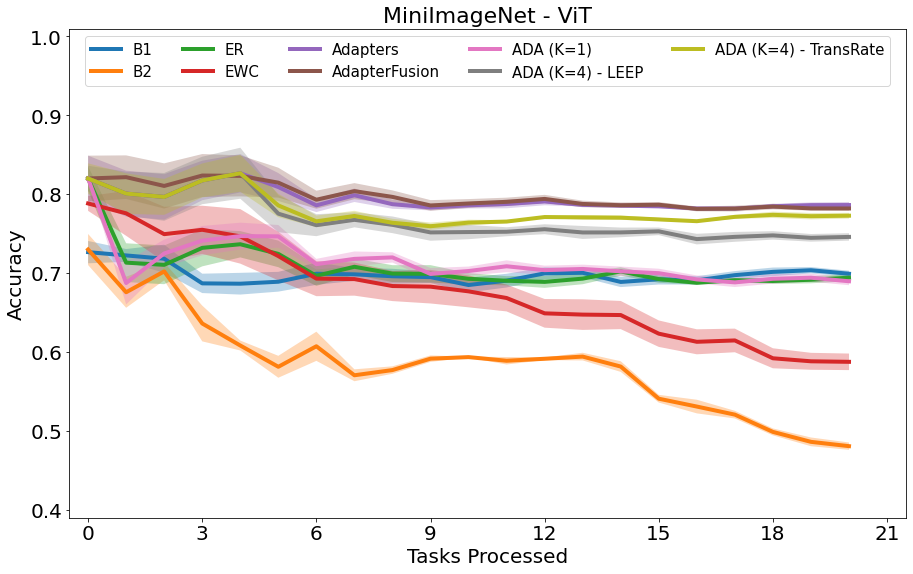}
\end{minipage}
\begin{minipage}{0.45\textwidth}
\includegraphics[width=\textwidth]{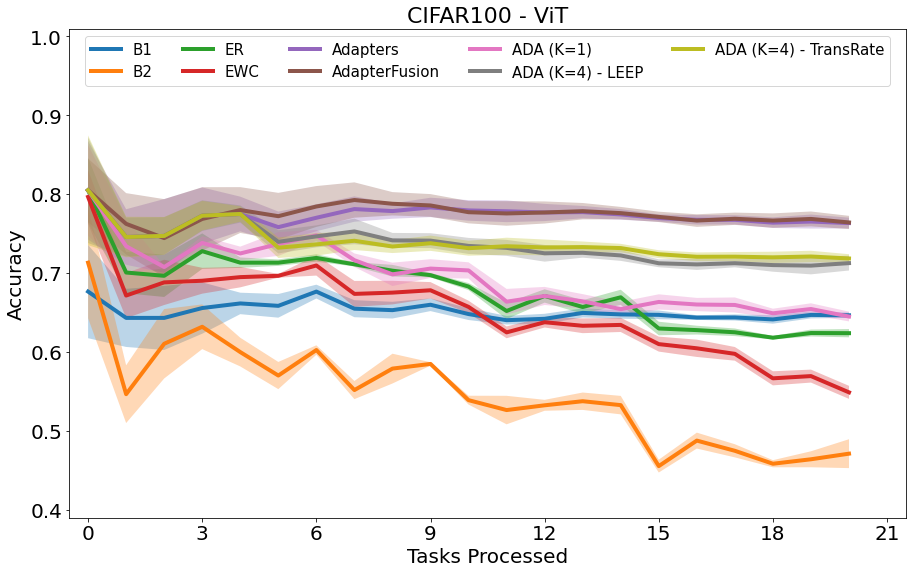}
\end{minipage}
\caption{Comparison between baselines and ADA with ViT model on MiniImageNet and CIFAR100.
Top figures shows the first scenario (binary) results, and bottom figures shows the second scenario (multi-class) results.
On the x-axis we report the number of tasks processed, on the y-axis we report the average accuracy measured 
on the test set of the tasks processed, shaded area shows standard deviation.}
\label{fig:compAccuracies}
\end{figure*}

\paragraph{Experimental Setup.} Both CIFAR100 and MiniImageNet consist of 60000 colour 
images in 100 classes, with 600 images per class.
We design two scenarios, in the first scenario each new task is a balanced binary classification problem where
the positive class is selected at random and the data points in the negative class are selected randomly
from the classes picked from the previous tasks. Each class can be selected to be
the positive class only once.
In the second scenario each task is a balanced multi-class classification problem with 5 classes. 
In both cases we provide the learner with 50 data points per class both at training and test time:
in the first scenario each task will have a training set of 250 data points and in the second case
of 100 data points. The total number of tasks is fixed to 20 for both scenarios.
We use \emph{Adam} as optimizer with the batch size of 8. For learning rate, we select best 
from $\lbrace 0.00005, 0.0001, 0.0005, 0.001\rbrace$ after observing the results on the first five tasks.

\paragraph{Adapter Architectures.} We use pre-trained models from HuggingFace Transformers~\cite{wolf2020transformers} as our base feature extractors. 
We ran experiments with ViT-B~\cite{dosovitskiy2020image}\footnote{\url{https://huggingface.co/google/vit-base-patch16-224}}
and DeiT-B~\cite{touvron2021training}\footnote{\url{https://dl.fbaipublicfiles.com/deit/deit_base_patch16_224-b5f2ef4d.pth}}.
Both models use 12 layers of transformers block with a hidden size of 768 and number of self-attention heads as 12 and has around 86 M trainable parameters.
We implemented the same adapter architecture for ViT and DeiT with AdapterBERT~\cite{houlsby2019parameter}. 
For all the adapter-based algorithms that we define in the following section, we use the same configuration for the adapters, setting the adapter hidden size to $48$.
With this setting, an adapter contains ~1.8 M parameters. We also train a head model for each task, that has the size of $\text{ViT-B}\times$ \emph{output size}.

\paragraph{Baselines.} We compare ADA with the following baselines.
1) \emph{Fine-tuning head model (B1)}: We freeze the pre-trained representation and only fine-tune the output layer of each classification task. 
The output layer is multiple-head binary classifier that we also use for the other methods.
2) \emph{Fine tuning the full model (B2)}: We fine-tune both the pre-trained representation and the output layer for each classification task.  
3) \emph{Adapters}~\citep{houlsby2019parameter}: We train and keep separate adapters for each classification task as well as the head models. 
4) \emph{AdapterFusion}~\citep{pfeiffer-etal-2021-adapterfusion}: It is a two stage learning algorithm that leverages knowledge from multiple tasks
by combining the representations from several task adapters in order to improve the performance on the target task. 
This follows exactly the solution depicted in Section~\ref{sec:ada}, Adapters.
5) \emph{Experience Replay (ER)}~\cite{rolnick2018experience}: ER is a commonly used baseline in Continual Learning 
that stores a subset of data for each task and then ``replays'' the old data together with the new one to avoid forgetting old concepts.
To make the results comparable, we use ER with same pre-trained network of the other algorithm and a single adapter being trained. Memory size is set to 500.
6) \emph{Elastic Weight Consolidation (EWC)}~\cite{kirkpatrick2017overcoming}: EWC is a regularization-based CL method that assumes that some weights
of the trained neural network are more important for previously learned tasks than others. During training of the neural network on a new task, 
changes to the weights of the network are made less likely the greater their importance. 
To estimate the importance of the network weights, EWC uses probabilistic mechanisms, in particular the Fisher information matrix.
We tune the regularization coefficient of EWC by grid search in $\lbrace 0,1,10,100,1000\rbrace$.
\\
In addition to these baselines, we use one special case of ADA with K=1, offering another reference point with the performance of ER and
helping to quantify the advantage of effective consolidation of adapters. All the results in this section are average of 5 runs.

%% Figure1. Accuracy comparison of methods
\begin{figure*}[ht!]
\centering
\begin{minipage}{0.45\textwidth}
\includegraphics[width=\textwidth]{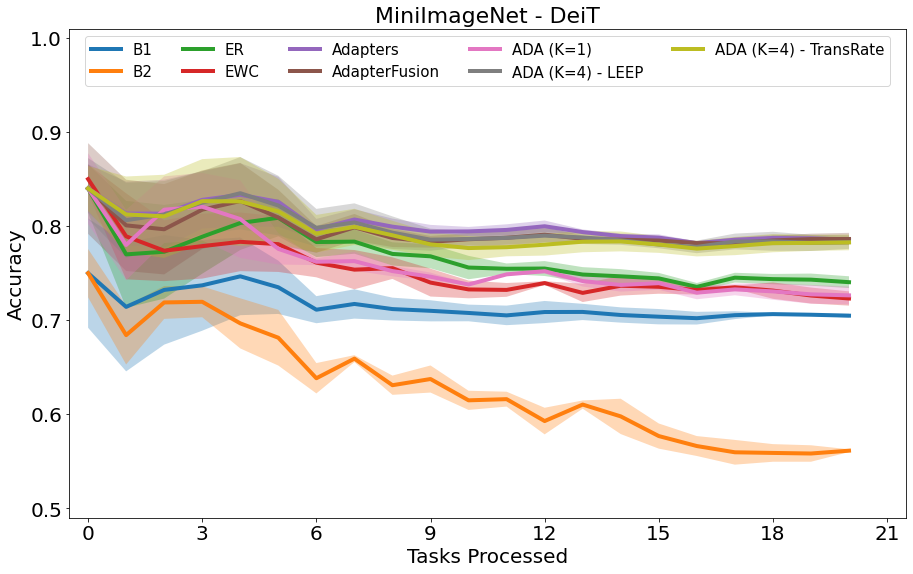}
\end{minipage}
\begin{minipage}{0.45\textwidth}
\includegraphics[width=\textwidth]{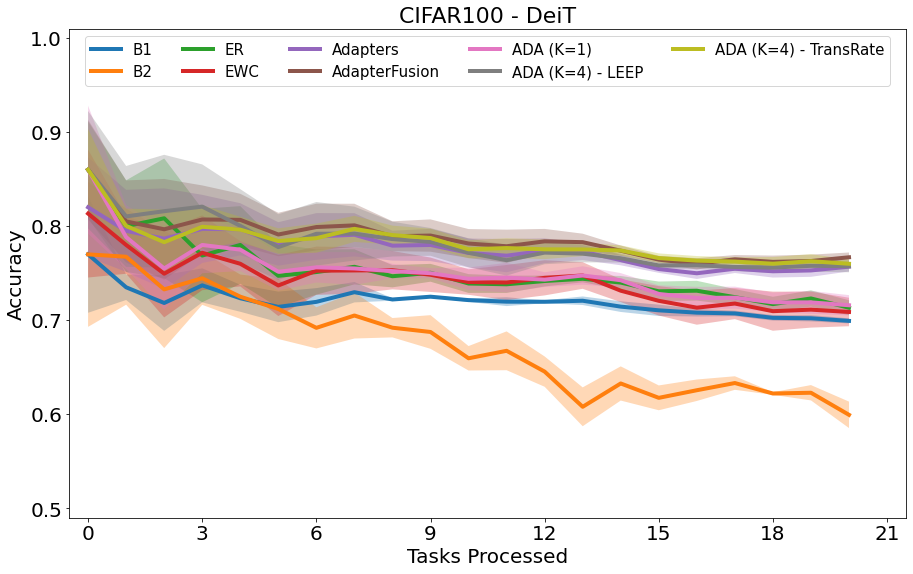}
\end{minipage}
\\
\centering
\begin{minipage}{0.45\textwidth}
\includegraphics[width=\textwidth]{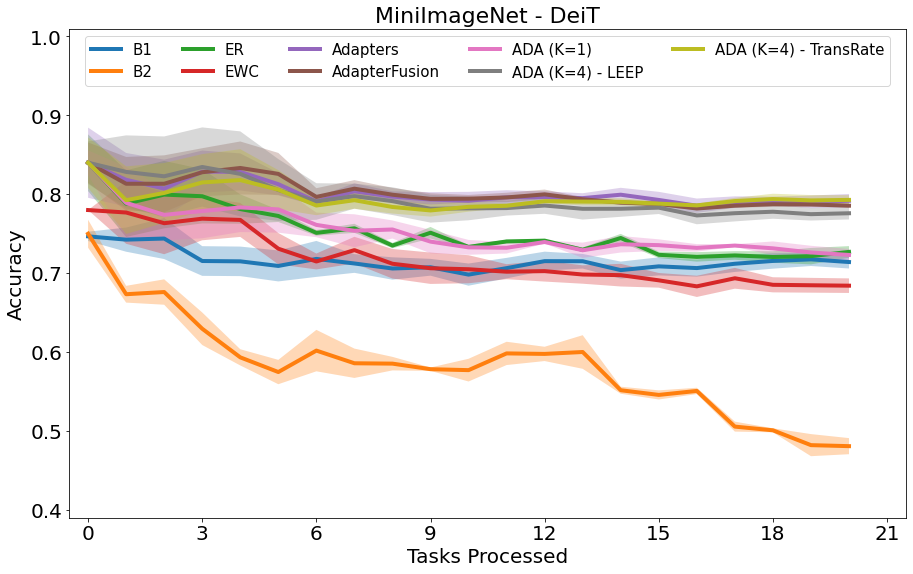}
\end{minipage}
\begin{minipage}{0.45\textwidth}
\includegraphics[width=\textwidth]{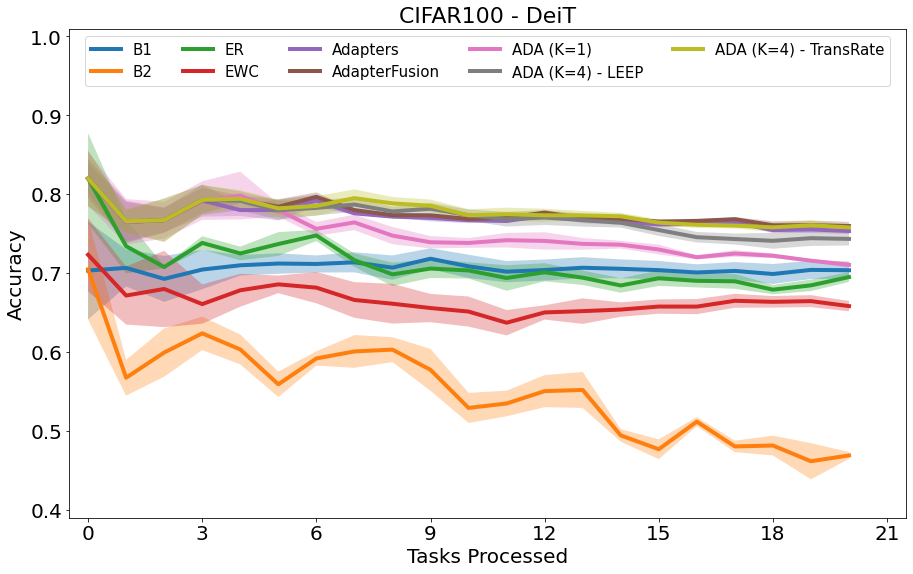}
\end{minipage}
\caption{Comparison between baselines and ADA with DeiT model on MiniImageNet and CIFAR100.
Top figures shows the first scenario (binary) results, and bottom figures shows the second scenario (multi-class) results.}
\label{fig:compAccuraciesDeiT}
\end{figure*}

\paragraph{Predictive performance.} Figure~\ref{fig:compAccuracies} shows the comparison of ADA and the baseline methods. 
It can be clearly seen that freezing all pre-trained model parameters, and fine-tuning only the head models (B1)
leaded to an inferior performance compared to other approaches. 
The main reason is that the head models have small amount of parameters to train and fine-tuning only the heads
suffers from under-fitting.
B2 performs well only for first 2-3 tasks, since we keep training the
complete model, it forgets the previously learned tasks very quickly. 
Adapters and AdapterFusion add ~1.8 M parameters for each task and
train these parameters with new task data, fixing them after training. 
There is no forgetting since the parameters are not shared among tasks
and the results on each dataset confirm this.
Although the careful tuning of regularization coefficient, EWC cannot handle CF,
especially for multi-class classification problem.
ADA with K=1 shows that distillation alone doesn't prevent forgetting. In almost all cases,
ER perform on-par with ADA K=1. providing evidence that a small amount of memory can actually improve performance 
compared to fine-tuning or regularization, but the improvement is limited and does not last as the number of tasks increases.
\\
ADA-LEEP and ADA-TransRate results with K=4 adapters show that selective consolidation of adapters significantly improve
the performance. For binary classification, their performance are on par with AdapterFusion while the number of model
parameters is significantly lower. For multi-class, their performance slightly declines after a certain number of tasks.
This is discussed in~\cite{ermis2022ada}, and the main reason is that the capacity of the adapter is exceeded.
Using a bigger adapter pool, or using larger adapters can solve the issue quickly, but to keep the comparison fair, we used the same
size adapters for each algorithm.
\\
To validate the interoperability of ADA to different models, we run the same experiments on DeiT model. Figure~\ref{fig:compAccuraciesDeiT}
demonstrates the same behaviour of algorithms with DeiT.

\paragraph{Memory consumption.} Table~\ref{table:numparam} reports the number of parameters used for baselines and ADA in our experiments.
We don't add the head size to the table, since it's very small: 768 parameters per binary head, ~15K parameters (6 KB) for 20 tasks,
3840 per multi-class head, ~75K parameters (30KB) for 20 tasks. Also they are same for all the methods. 
\begin{table}[H]
\begin{minipage}[b]{1.0\hsize}\centering
\scalebox{0.75}{
\begin{tabular}{l | l | l | l | l }
           & \multicolumn{4}{c }{Fine-Tuning (B1, B2) and EWC}  \\  \hline
           & Trainable  & Inference  & Total    &  Total (Size) \\ \hline \hline
Task = $\lbrace 1,10,20\rbrace$   & 86 M    &  86 M   &  86 M   &  344 MB   
\end{tabular}
}
\end{minipage}
\\
\begin{minipage}[b]{1.0\hsize}\centering
\vspace*{3mm}
\scalebox{0.75}{
\begin{tabular}{l | l | l | l | l }
           & \multicolumn{4}{c }{Adapters \& AdapterFusion}                       \\  \hline
           & Trainable  & Inference   & Total          & Total (Size)             \\ \hline \hline
Task = 1 \hspace*{2mm}  & 1.8 M       &  87.8 M  &  87.8 M  & 351.2 MB          \\ \hline
Task = 10  & 1.8 M  &  86 + (F$\times$1.8) M  &  104 M   & 416 MB                \\ \hline
Task = 20  & 1.8 M  &  86 + (F$\times$1.8) M  &  122 M   & 488 MB                 \\ \hline
\end{tabular}
}
\end{minipage}
\\
\begin{minipage}[b]{1.0\hsize}\centering
\vspace*{3mm}
\scalebox{0.67}{
\begin{tabular}{l | l | l | l | l }
           & \multicolumn{4}{c }{ADA}                                       \\  \hline
           & Trainable  & Inference   & Total     & Total (Size)             \\ \hline \hline
Task = 1   & 1.8 M      & 102.8 M     & 102.8 M   & 452.2 MB   \\ \hline
Task = $\lbrace 10,20\rbrace$   &  2$\times$1.8 M &  102.8 M   &  101 + (K+1)$\times$1.8 M  &  445 + (K+1)$\times$7.2 MB
\end{tabular}
}
\end{minipage}
\caption{The number of all parameters and those used for training and inference as well as the model size of methods for ViT-B (Same for Deit-B).
$K$ is the number of adapters in the pool, and $F$ is the number of fused adapters (it is between 2 and number of tasks). For the \emph{Adapters} $F=1$.
For ER, for Task = $\lbrace 1,10,20\rbrace$, it is same with ADA Task=1.}
\label{table:numparam}
\end{table}
These results make clear that ADA is significantly more efficient in terms of memory usage. It can achieve 
predictive performance similar to the one of Adapters and AdapterFusion while requiring significantly less model parameters.
ADA stores only 5 Adapters (K=4 adapters in the pool, and one adapter for new task), against the 20 required by AdapterFusion.

\paragraph{Similarities and differences with the NLP experiments}
In~\cite{ermis2022ada}, they address incremental binary text classification. In this paper, we further investigated the multi-class classification where
we observe a total of 100 classes. Almost all algorithms show similar behaviour with NLP experiments. One interesting outcome is the performance of full
fine-tuning (B2) performed better with image transformers. The accuracy declines more significantly on NLP datasets with NLP transformers.
Besides LEEP works better with image datasets as it was originally proposed for CV tasks. There is mostly no difference between ADA-LEEP and ADA-TransRate, 
or the difference is very small. But in~\cite{ermis2022ada}, the difference is noticable.

\section{Conclusion}
\label{sec:conc}

In this work we show that the usage of Adapters in combination with Transformers 
for continual CV problems. In particular, utilizing ad-hoc algorithms, such as \al,
can give a strong result in terms of predictive performance with constant parameter
increase.
Vision transformers have been known to have a tendency to overfit training datasets, 
consequently leading to poor predictive performance in small data regimes,
however~\cite{park2022vision} just showed that this claim is poorly supported, 
explains the nature of multi-head self-attentions and shows that ViT does not
overfit even on smaller datasets. This work is encouraging for the future studies 
in CV with vision transformers, and any development in that field will positively
impact CL research with Transformers. 

There are some aspects of our results which we would like to further investigate 
in the future. For instance, in this work we adopted the same Adapters structure
leverage in the NLP domain and, while that gave good results, there is the possibility
that it is a suboptimal choice. Also, we tested ADA with a fixed number of Adapters
but it is easy to observe that the number of Adapters could be controlled
by the algorithm itself, for example leveraging heuristics 
based on thresholding the LEEP or TransRate scores.

%%%%%%%%% REFERENCES
{\small
\bibliographystyle{ieee_fullname}
\bibliography{paper}
}

\end{document}